\documentclass[journal]{IEEEtran}

\usepackage{multibib}
\usepackage{graphicx}
\usepackage{mathptmx} 
\usepackage{newtxtext}
\usepackage{textcomp}
\usepackage[varg,bigdelims]{newtxmath}

\usepackage{enumitem}

\usepackage{amsmath}
\usepackage{dsfont, url}
\usepackage{amsthm}
\usepackage{mathtools}
\usepackage[usenames, dvipsnames]{xcolor}
\usepackage[%
colorlinks={true},%
citecolor={Blue},%
urlcolor={PineGreen},%
linkcolor={Sepia}%
]{hyperref}
\usepackage{float}
\usepackage{newfloat}
\usepackage[skip=1pt]{caption}

\usepackage{amsmath} 
\usepackage{algorithm}
\usepackage[noend]{algpseudocode}
\usepackage{todonotes}
\usepackage{pgf,tikz}
\usepackage{adjustbox}
\usepackage{verbatim} 
\usepackage{xargs}
\usepackage{cite}
\usepackage{multirow}

\usetikzlibrary{plotmarks}
\usetikzlibrary{chains}
\usepackage{pgfplots}
\pgfplotsset{compat=newest}
\pgfplotsset{plot coordinates/math parser=false}
\newlength\figureheight
\newlength\figurewidth
\usetikzlibrary{shapes,arrows}
\usetikzlibrary{positioning, calc, fit}
\usetikzlibrary{decorations,decorations.pathmorphing,decorations.pathreplacing}
\usetikzlibrary{shapes.misc}
\usetikzlibrary{shadows,trees}

\DeclareMathOperator*{\sbjto}{s.\ t.\ }

\DeclareMathOperator{\trace}{tr}
\DeclareMathOperator{\proj}{Proj}

\newcommand{\R}{\mathds{R}}
\newcommand{\Nz}{\mathds{N}}
\newcommand{\N}{\mathds{Z}_{+}}
\newcommand{\z}{\mathds{Z}}

\newcommand{\bmat}[1]{\begin{bmatrix}#1\end{bmatrix}}

\newcommand{\norm}[1]{\left\|#1\right\|}
\newcommand{\secref}[1]{\S \ref{#1}}

\renewcommand{\transp}{^\top}

\newcommand{\zeros}{\mathbf{0}}

\newcommand{\st}{x}

\newcommand{\control}{u}
\newcommand{\controlset}{\mathds{U}}

\newcommand{\Let}{\coloneqq}

\newtheorem{assumption}{Assumption}

\newtheorem{theorem}{Theorem}
\newtheorem{lemma}{Lemma}

\newtheorem{definition}{Definition}

\newtheorem{pstatement}{Problem Statement}

\allowdisplaybreaks

\begin{document}
	
	\title{Unmatched uncertainty mitigation through neural network supported model predictive control}
	
 \author{Mateus V. Gasparino, Prabhat K. Mishra,~\IEEEmembership{Member,~IEEE,} and~Girish Chowdhary,~\IEEEmembership{Senior Member,~IEEE}%
		\thanks{M. V. Gasparino and G. Chowdhary are with Computer Science Department, University of Illinois at Urbana Champaign (UIUC), 
			USA. \tt\{mvalve2,girishc\}@illinois.edu}
   \thanks{P. K. Mishra is with Chemical Engineering Department, Massachusetts Institute of Technology (MIT), Cambridge, USA. \tt{pkmishra@mit.edu}} 		
	}

	\maketitle
	
\begin{abstract}
This paper presents a deep learning based model predictive control (MPC) algorithm for systems with unmatched and bounded state-action dependent uncertainties of unknown structure. We utilize a deep neural network (DNN) as an oracle in the underlying optimization problem of learning based MPC (LBMPC) to estimate unmatched uncertainties. Generally, non-parametric oracles such as DNN are considered difficult to employ with LBMPC due to the technical difficulties associated with estimation of their coefficients in real time. We employ a dual-timescale adaptation mechanism, where the weights of the last layer of the neural network are updated in real time while the inner layers are trained on a slower timescale using the training data collected online and selectively stored in a buffer. Our results are validated through a numerical experiment on the compression system model of jet engine. These results indicate that the proposed approach is implementable in real time and carries the theoretical guarantees of LBMPC.
\end{abstract}


\section{Introduction}
\par Machine learning and Model Predictive Control (MPC) complement each other by compensating drawbacks of each other and making their combination useful for safety critical applications. We refer readers to \cite{survey_LMPC, Survey_Schoellig} for excellent surveys on safe learning and robotics. 
\par Learning based Model Predictive Control (LBMPC) \cite{lbmpc_linear} became popular due to its improvement over linear MPC in terms of transient response and overshoot with slight expense in processing time \cite{lbmpc_quadcopter}. Several interesting applications such as autonomous driving \cite{zieger_CDC_2022, lmbpc_copilot}, heat, ventilation and air-conditioning systems \cite{lbmpc_HVAC}, quad-copter \cite{lbmpc_quadcopter}, formation control \cite{lbmpc_formation_control}, atmospheric pressure plasma jets \cite{lbmpc_Mesbah}, air-borne wind energy systems \cite{Ecket_ACC_2022} etc., contributed in theoretical and practical advancements. 

\par Due to the significant success of and progress in deep learning techniques, this is tempting to use a neural network as an oracle in LBMPC. However, boundedness and differentiability of the oracle is required to hold the results of \cite{lbmpc_linear}. Even though a bounded and differentiable neural network can be constructed, the estimation of their weights in real time is generally difficult \cite{vapnik1999}. Since training of DNN is a time demanding process, real time implementation of DNN supported LBMPC is challenging. 
\par In this article, we demonstrate that the real time implementation of neural network based oracle is not only possible but also computationally efficient by two time scale training. In \cite{Joshi_DMRAC, PG20}, the training of the output layer and the hidden layers are separated in which the output layer is trained online through adaptation by considering the recently trained hidden layers as a feature basis function and the hidden layers are trained on a parallel machine by keeping the recently updated weights of the output layer fixed. We refer readers to excellent surveys on transfer learning \cite{transfer_learning_2010, transfer_learning_2020}. This method of two time scale training allows us to try different methods of training the output layer and hidden layers independently. In addition, different architectures such as Recurrent Neural Network and Convolutional Neural Network are also supported for the hidden layer. The theoretical guarantees mostly depend on last activation layer and the training mechanism of the linear output layer. The above methods \cite{Joshi_DMRAC, PG20} are limited only to those uncertainties that enter into the dynamics through control channel. The present article extends the results of \cite{Joshi_DMRAC, PG20} by generalizing the class of uncertainties. Our approach improves the performance of \cite{lbmpc_linear} by replacing the L2NW estimator by a DNN. Since PyTorch is popular for DNN implementation and CasAdi for MPC, we address some non-trivial issues related to their interface also.
\begin{figure}
\centering
\includegraphics[width=0.9\linewidth]{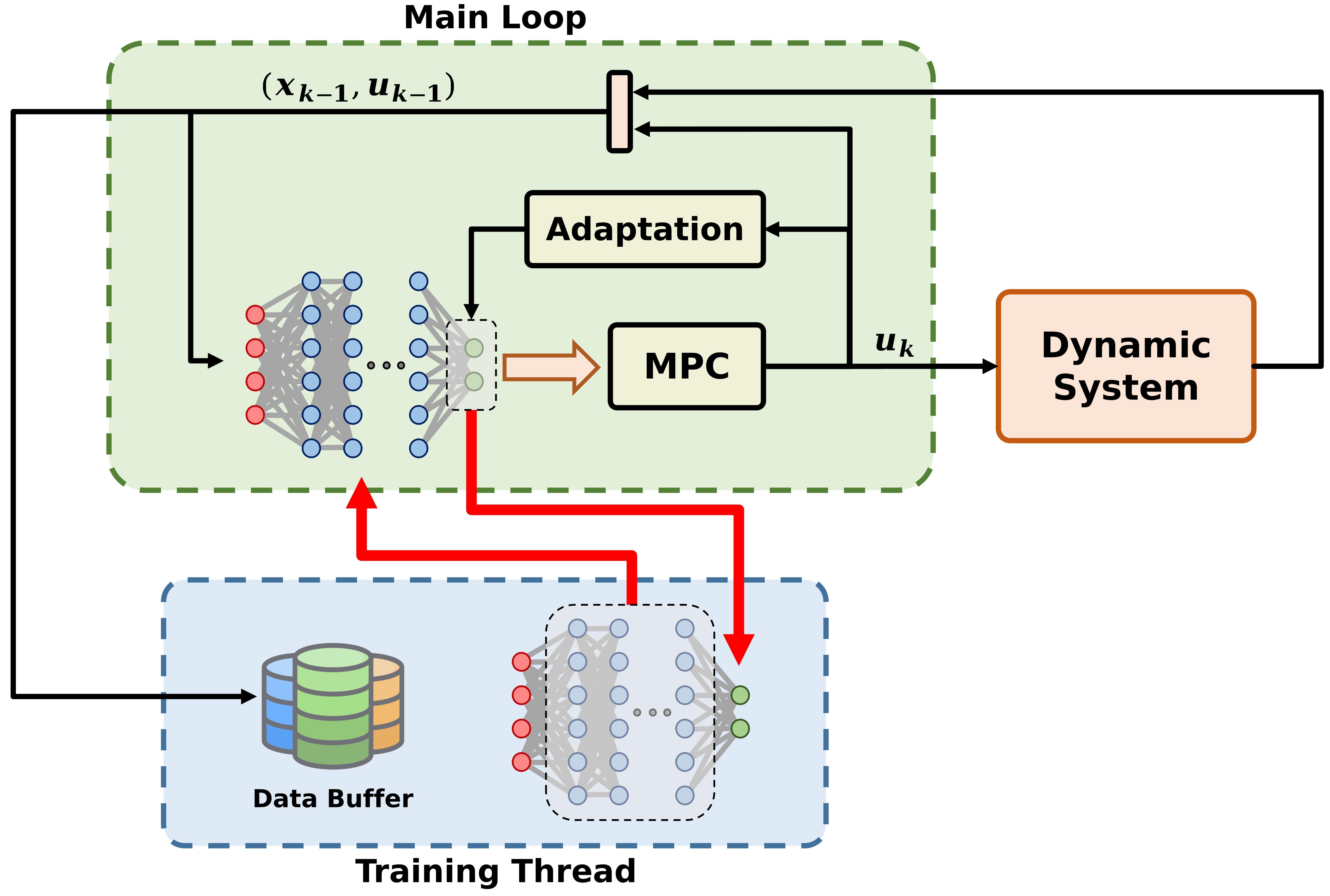}
	\caption{The neural network on the main loop with fixed hidden layers is connected with MPC and transmits the weights of output layer at time sequence $(T_k)_{k\in \N}$ to the neural network on the training thread. This neural network (on the training thread) transmits the weights of hidden layer to the neural network (on the main loop) at time instants $(t_j)_{j\in \N}$ after its $j^{\text{th}}$ training.}
	\label{fig:schematic}
\end{figure}
\par This article is organized as follows. The problem statement is given in \secref{s:setup}. The architecture of neural network and its training mechanism are explained in \secref{s:dnn}. LBMPC and its properties are presented in \secref{s:MPC} and \secref{s:Properties}, respectively. We validate our results in \secref{s:experiment} and conclude in \secref{s:epilogue}.  	
\par We let $\R$ denote the set of real numbers, $\Nz$ the set of non-negative integers and $\N$ the set of positive integers. For a given vector $v$ and positive (semi)-definite matrix $M \succeq \zeros$, $\norm{v}_M^2$ is used to denote $v \transp M v$. For a given matrix $A$, the trace, the largest eigenvalue and pseudo-inverse are denoted by $\trace(A)$, $\lambda_{\max}(A)$ and $A^{\dagger}$, respectively. By notation $\norm{A}$ and $\norm{A}_{\infty}$, we mean the $2-$norm and $\infty-$norm when $A$ is a vector; induced $2-$norm and $\infty-$norm when $A$ is a matrix, respectively. A vector or a matrix with all entries $0$ is represented by $\zeros$ and $I$ is the identity matrix of appropriate dimensions. We let $M^{(i)}$ denote the $i^{\text{th}}$ column of a given matrix $M$.
\section{Problem setup}\label{s:setup}

Let us consider a discrete time dynamical system 

\begin{equation}\label{e:system}
    \st_{t+1} = A\st_t + B\control_t + h(\st_t, \control_t); \quad t \in \Nz,
\end{equation}

\begin{enumerate}[leftmargin = *, nosep, label=(1-\alph*), widest = b]
	\item \label{e:constraints} $\st_t \in \mathcal{X} \subset \R^d$, $ \control_t \in \controlset \subset \R^m; \quad d,m \in \N$,
	\item $h:\mathcal{X}\times \controlset \rightarrow \R^d$ is a continuous and (possibly) nonlinear function, which represents state-action dependent unmatched uncertainty (or modeling error),
	\item \label{as: bounds_hg} $h(\st, \control) \in \mathds{W}$ for every $\st \in \mathcal{X}$ and $\control \in \controlset$,
	\item $\mathcal{X}, \controlset$ and $\mathds{W}$ are polytopes,
	\item the matrix pair $(A,B)$ is stabilizable. 
\end{enumerate}

The matrices $A$ and $B$ represent our domain knowledge or prior knowledge about the system dynamics, and the continuous function $h$ represents the unknown component of the system dynamics\footnote{For example, when a non-linear dynamics is linearized by the matrix pair $(A,B)$, $h$ represents the linearization error.}. 
LBMPC \cite{lbmpc_linear} has been developed to improve the closed-loop performance of tube-based robust MPC by modifying the cost function in the underlying optimization problem. The cost function is modified by learning (or estimating) the unknown function $h$ with the help of data. In this article, we address issues associated with the use of DNN as an estimator of $h$. Our main focus is to use DNN in such a way that it can be implemented in real time on a hardware with limited computational power. The general problem description of this article is as follows:
	
\begin{pstatement}\rm{
    Present a stabilizing, robust and real-time implementable control framework for \eqref{e:system}, which respects physical constraints \ref{e:constraints}, optimizes a given performance index, and reduces the effect of unmatched uncertainties by using a trainable DNN.}   
\end{pstatement}

\section{Deep neural network}\label{s:dnn}

Any continuous function $h$ on a compact set $\mathcal{X}\times \controlset$ can be approximated by a multi-layer network with number of layers $L \geqslant 2$ such that   

\begin{equation}
	h(x, u) = W_L\transp \psi_L \left[ W_{L-1}\transp \psi_{L-1} \left[ \cdots \left [ \psi_1(x,u)\right] \right] \right] + \varepsilon^{\ast}(x,u),
\end{equation} 

\noindent where $x \in \mathcal{X}, u \in \controlset,  \psi_i, W_i$ for $i=1, \ldots , L$, are the activation functions in the $i^{\text{th}}$ layer and the corresponding ideal weights, respectively \cite[\S 7.1]{Lewis_NN_99}. We can represent $h(\st_t, \control_t)$ with the help of a neural network with a desired accuracy. 
	
Let us define $\phi^{\ast}(x,u) \Let \psi_L \left[ W_{L-1}\transp \psi_{L-1} \left[ \cdots \left[ \psi_1(x,u)\right] \right] \right]$ and $W^{\ast} \Let W_{L} $, then	 
	
\begin{equation}\label{e:NN_structure}
	h(\st_t, \control_t) = W^{\ast \top}\phi^{\ast}(\st_t,\control_t) + \varepsilon^\ast(\st_t, \control_t),
\end{equation} 
	
\noindent where $W^{\ast} \in \R^{(n_L + 1) \times d}$ denotes the weights of the output layer. There are $n_L$ number of neurons in the last hidden layer. The first row of $W^{\ast}$ represents the bias term in the output layer and the first element of $\phi^\ast \in \R^{n_L+1}$ is $1$.

\begin{figure}
\begin{center} 
	\includegraphics[width=0.9\linewidth]{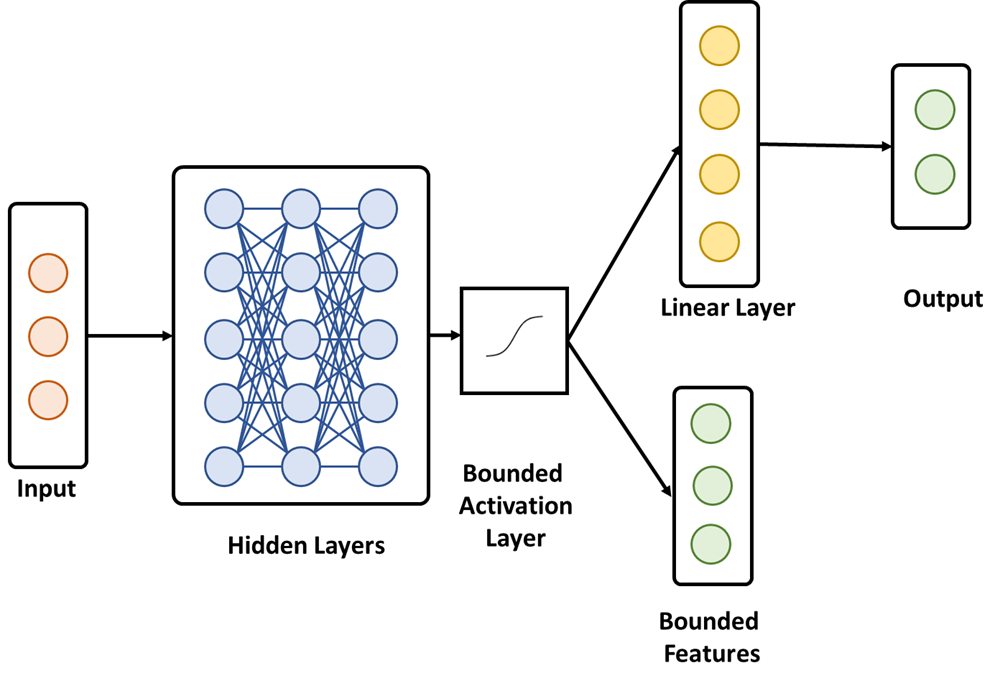}
	\caption{The neural network architecture can use an arbitrary number of layers and neurons. The input layer has the same number of neurons as the experience data. The output layer has as many neurons as system states. The hidden layers can have any architecture provided suitable adjustments are made to facilitate their training.}
	\label{fig:neural-network}
\end{center}
\end{figure}

The major challenge is associated with real-time implementation of DNN because its training takes much more time than the sampling interval of fast hardware like quad-copter. Therefore, we address this issue with the help of two DNN's as shown in Fig.\ \ref{fig:schematic}. 
Both DNNs have same architecture as in Fig.\ \ref{fig:neural-network}. 
The DNN in the main loop is located on the main machine and the other DNN is located on some secondary (or remote) machine. We update the weights of the output layer on the main machine in real time at each time instant with the help of a weight update law while keeping the weights of hidden layers fixed. The hidden layers are trained on a secondary machine by using the approach \cite{DMRAC} in which the weights of the output layer are copied from the main machine at the start of the training and remain fixed during the training. Once the training of DNN on a secondary machine is complete, new weights of hidden layers are updated on the main machine and remain fixed until the new set of weights are again obtained from the secondary machine. Training details about the output layer and hidden layers are provided in \secref{s:outer layer training} and \secref{s:inner layer training}, respectively.

\par Let $(t_j)_{j \in \N}$ represent an increasing time sequence of instants when weights of hidden layers are updated on main machine. Therefore, we get the following expression:

\begin{equation}\label{e:NN_jth_instant}
	h(\st_t, \control_t) = W^{\ast \top}\phi_j(\st_t,\control_t) + \varepsilon_j(\st_t, \control_t),
\end{equation}
where $\varepsilon_j(\st_t, \control_t) = W^{\ast \top} \left( \phi^{\ast}(\st_t, \control_t) - \phi_j(\st_t, \control_t) \right) + \varepsilon^\ast(\st_t, \control_t)$.
\noindent We can assume that $\norm{\phi_j(x,u)}$ will be bounded for each $j$,  $u \in \R^m$ and $x \in \R^d$ due to the presence of bounded activation layer in Fig. \ref{fig:neural-network} consisting bounded neurons, i.\ e.\ sigmoidal, tanh, etc. Since $h$ is bounded due to \ref{as: bounds_hg}, we can assume that ideal weights $W^\ast$ in the output layer are also bounded. We make the following assumption: 

\begin{assumption}\label{as:bounds_uncertainty}
    \rm{
    	There exist $\bar{W}_{i} > 0$ for $i = 1, \ldots , d$, and $\sigma, \bar{\varepsilon} > 0$ such that $\norm{W^{\ast (i)}} \leqslant \bar{W}_{i}$, for $i = 1, \ldots , d$, and $\norm{\phi_j(x, u)} \leqslant \sigma$ for every $x \in \mathcal{X}, u \in \controlset$ and $j \in \Nz$.}   
\end{assumption}

\noindent The above assumption is standard in literature \cite{HHN_08, NN_15, Joshi_DMRAC}. If the neural network is not minimal then the ideal weights may not be unique. However, for the neural-adaptive controller design only the existence of ideal weights is assumed, which is always guaranteed when $h$ is a continuous function on a compact set \cite[\S 7.1]{Lewis_NN_99}. A priori knowledge about the bounds on the ideal weights $W^\ast$ of the output layer is useful to avoid parameter drift phenomenon.  

\par At $t_0 = 0$, weights of DNN on both machines are randomly initialized with desired bound on the output layer weights as per the Assumption \ref{as:bounds_uncertainty}. Therefore, for $j \in \Nz$, we have
\begin{equation}\label{e:adaptive_control}
	\hat{h}_t(\st_t, \control_t) \Let K_t\transp \phi_j(\st_t, \control_t) \text{ for } t \in \{t_j, t_j +1, \ldots , t_{j+1}-1 \}.  
\end{equation}
	
\noindent We update $K_t$ in an unsupervised manner on main machine while collecting the training data for the DNN on secondary machine. In the next subsections we provide the relevant details of the training of DNN.

\subsection{Adaptive learning of $W^\ast$ on the main machine}\label{s:outer layer training}

For $t \in \{t_j, t_j +1, \ldots , t_{j+1}-1 \} $, the output of DNN is given by \eqref{e:adaptive_control} and the bounded features are given by $\phi_j(x_t,u_t)$ at time $t+1$ \footnote{This is important for implementation to note that $u_t$ is computed at time $t$ but we need to use $\phi_j(x_t,u_t)$ and $\hat{h}(x_t,u_t)$ at time $t+1$.}. We get the estimated state 
	\begin{equation}
	\hat{\st}_{t+1} = A\st_t + B\control_t +\hat{h}_t(\st_t, \control_t).
	\end{equation}
 We can compute the error $\tilde{\st}_{t+1} = \hat{\st}_{t+1} - \st_{t+1}$. 
For the online training of the output layer, we employ the projection based robust weight update law and refer readers to \cite[Chapter 10]{Landau_Karimi_11} for other methods. For a given learning rate $0< \gamma < 1$, we first modify $K_t$ by taking $\tilde{\st}_{t+1}$ into account to get $\bar{K}_{t+1}$ and then project $\bar{K}_{t+1}$ to ensure its boundedness as follows:
	\begin{equation}\label{e:weight_update_law}
	\begin{aligned}
	& \bar{K}_{t+1} = K_t - \gamma \frac{\phi_j(\st_t, \control_t)}{\norm{\phi_j(\st_t, \control_t)}^2}  \tilde{\st}_{t+1} \transp ,\\
	& K_{t+1}^{(i)} = \proj \bar{K}_{t+1}^{(i)} = \begin{cases} \bar{K}_{t+1}^{(i)} & \text{ if } \norm{\bar{K}_{t+1}^{(i)}} \leqslant \bar{W}_{i} \\
	\frac{\bar{W}_{i}}{\norm{\bar{K}_{t+1}^{(i)}}} \bar{K}_{t+1}^{(i)} & \text{ otherwise. } \end{cases}.
	\end{aligned}
	\end{equation}	
The implications of \eqref{e:weight_update_law} are discussed in \secref{s:Properties}. 

\subsection{Supervised learning of $\phi^\ast$ on a secondary machine}\label{s:inner layer training}

In this section, the training data and the loss function required for the supervised training of the DNN on secondary machine are explained.
For a given state-action pair $(\st_t, \control_t)$ as input at time $t$, the label $h(\st_t, \control_t)$ is computed at time $t+1$ by the relation:
\begin{equation}
    h(\st_t, \control_t) = \st_{t+1} - A\st_t - B\control_t.
\end{equation}
These pairs $(x,u)$ and corresponding labels $h(x,u)$ are stored in a training buffer until the buffer is full. In particular, a full buffer consists of $\left( (x^i,u^i), h(x^i,u^i) \right)$ for $i=1, \ldots, p_{\max}$. 

Once the buffer is full, new data is stored by replacing some old data. Several methods are proposed in the literature to increase the richness of the training buffer; see \cite{cautious_MPC, input_feature, Morari_GP} and references therein. We refer readers to \cite{experience_selection} for different methods of designing replay buffer.

Let $(T_k)_{k\in \N}$ be an increasing time sequence.
At time $T_k$, the weight of output layer of the primary neural network are copied in the secondary neural network. During the training, the output layer weights in the secondary network remain fixed. Therefore, we are interested in finding the weights $W_{1:L-1} \Let W_1, \ldots, W_{L-1}$ which minimize the following cost for a given input $(x,u)$ and label $h(x,u)$:

\begin{align*}
	& \ell\left( (x,u), W_{1:L-1} \right) \Let \\
	 & \quad \quad \norm{h(x,u) - K_{T_k}\transp \psi_L \left[ W_{L-1}\transp \psi_{L-1} \left[ \cdots \left [ \psi_1(x,u)\right] \right] \right]}^2. 
\end{align*}
	
Let $M$ represent the number of training samples and $\mathcal{D}_k \Let \left( (x^i,u^i), h(x^i,u^i) \right)_{i=1}^M$ is training data consisting $M$ data points randomly sampled from  the buffer for the $k^{\text{th}}$ training. The following loss function is considered for the training of DNN:

\begin{equation*}
	\mathcal{L}(\mathcal{D}_k, W_{1:L-1}) = \frac{1}{M} \sum_{i=0}^M \ell\left( (x^i,u^i), W_{1:L-1} \right).
\end{equation*}

\section{Model predictive controller}\label{s:MPC}
We first fix an optimization horizon $N\in \N$. Let $x^r, u^r$ be a reference (or equilibrium) state-action pair. For given positive definite matrices $Q,R\succ 0$, $P\succ 0$ is the solution of the following Lyapunov equation:
\begin{equation}
    (A+BK)\transp P (A+BK) - P = -(Q+K\transp R K),
\end{equation}
where $K$ is such that $A+BK$ is Schur stable. We define the following cost
\begin{align*}
	\psi(z_{0:N+1},v_{0:N}) &\Let \norm{z_N - x^r}_P^2 + \sum_{i=0}^{N-1} \norm{z_i - x^r}_Q^2 + \norm{v_i - u^r}_R^2.
\end{align*} 
Let us define $R_{i+1} = (A+BK)R_i \oplus \mathds{W}$ with $R_0 = \{0 \}$. For $i=0,\ldots, N-1$, we impose the following constraints:
\begin{equation}\label{e:tightened_constraints}
	\begin{aligned}
		& \bar{z}_{i+1} = A\bar{z}_i + Bv_i \\
		& \bar{z}_i \in \mathcal{X}\ominus R_i, \quad  
		v_i \in \controlset \ominus KR_i \\
        & \bar{z}_N \in \Omega \ominus R_N,
	\end{aligned}
\end{equation}
where $\Omega$ is a disturbance invariant set. 
At each time $t$, we measure the state $\st_t$ of the system \eqref{e:system} and solve the following optimization problem:
		\begin{equation}\label{e:LBMPC}
		\begin{aligned}%
		\min_{(c_{i})_{i=0}^{N-1} } & \quad \psi(z_{0:N+1},v_{0:N}) \\
		\sbjto \ & \quad z_{0} = \bar{z}_0 = \st_t \\
		& \quad v_i = K\bar{z}_i + c_i \text{ for } i \in \z_{[0,N-1]}\\
		&\quad  z_{i+1} = Az_i + Bv_{i} + \hat{h}_t(z_{i}, v_{i}) \text{ for } i \in \z_{[0,N-1]}    \\
		& \quad \eqref{e:tightened_constraints}.
		\end{aligned}	
		\end{equation}
		By solving the above problem, we get $v_0 = K\st_t + c_0$. We set $\control_t = v_0$ and apply $\control_t$ to the system \eqref{e:system}.

\section{Properties of LBMPC}\label{s:Properties}
For the purpose of analysis, we define $\tilde{K}_t \Let K_t - W^\ast$, $\tilde{x}_{t+1} \Let \hat{x}_{t+1} - \st_{t+1} = \hat{h}_{t}(\st_{t}, \control_{t}) - h(\st_{t}, \control_{t}) = \tilde{K}_t\transp \phi_j(\st_t,\control_t) - \varepsilon_j(\st_t,\control_t) $ and $\bar{W} \Let \sum_{i=1}^m \bar{W}_i^2$. We recall the following definition:
	\begin{definition}[\cite{Ioannou_Fidan_tutorial}, page 117]\label{def:small_disturbance}
		\rm{
			The vector sequence $(s_t)_{t \in \Nz}$ is called $\mu$ small in mean square sense if it satisfies $\sum_{t=k}^{k+N-1}\norm{s_t}^2 \leqslant Nc_0 \mu + c_0^{\prime}$ for all $k \in \N$, a given constant $\mu \geqslant 0$ and some $N \in \N$, where $c_0, c_0^{\prime} \geqslant 0$. } 
	\end{definition}
 We make the following assumption:
 \begin{assumption}\label{as:bounded_reconstruction_error}
 \rm{
 There exists $\bar{\varepsilon}> 0$ such that 
 \begin{equation*}
 \norm{\varepsilon_j(x,u)} \leqslant \bar{\varepsilon} \text{ for each } (x,u) \in \mathcal{X} \times \controlset \text{ and } j \in \Nz.
 \end{equation*}
 }
 \end{assumption}
 Since $h$ is bounded due to \ref{as: bounds_hg}, $W^\ast, \phi_j$ due to Assumption \ref{as:bounds_uncertainty} and $K_t$ due to \eqref{e:weight_update_law}, the above assumption is trivially satisfied. We have the following result that says that the estimation error $\tilde{x}_{t+1} = \hat{h}_t(x_t,u_t) - h(x_t,u_t)$ is small in mean square sense. 
	\begin{lemma}\label{lem:adaptation}
		Consider the dynamical system \eqref{e:system}, weight update law \eqref{e:weight_update_law}. Let the assumptions \ref{as:bounds_uncertainty} and \ref{as:bounded_reconstruction_error} hold. Let us define $V_a(K_t) \Let \frac{1}{\gamma}\trace(\tilde{K}_t\transp\tilde{K}_t)$. Then for all $t$,
		\begin{enumerate}[label={(\rm \roman*)}, leftmargin=*, widest=3, align=left, start=1]
			\item \label{lem:bound_Va} $V_a(K_t) \leqslant \frac{4}{\gamma} \bar{W}$, 
			\item \label{lem:drift_Va} $V_a(K_{t+1}) - V_a(K_t)  \leqslant - \frac{1- \gamma}{\sigma^2} \norm{\tilde{\st}_{t+1}}^2 +  \norm{\varepsilon_j(\st_t,\control_t)}^2$,
			\item \label{lem:small_tildeu} $\tilde{\st}_t$ is $\bar{\varepsilon}^2$ small in mean square sense with $c_0 = \frac{\sigma^2}{1-\gamma }$ and $c_0^{\prime} = \frac{4c_0}{\gamma}\bar{W}$ as per the Definition \ref{def:small_disturbance}.	
		\end{enumerate}			
	\end{lemma}
	
\begin{proof}
    (Proof of Lemma \ref{lem:adaptation})
    \begin{enumerate}[label={(\rm \roman*)}, leftmargin=*, widest=3, align=left, start=1]
	
	\item Since $V_a(K_t) = \frac{1}{\gamma} \trace(\tilde{K}_t\transp \tilde{K}_t) = \frac{1}{\gamma} \sum_{i=1}^m \norm{K_t^{(i)} - W^{\ast (i)}}^2 \leqslant \frac{2}{\gamma} \sum_{i=1}^m  \norm{K_t^{(i)}}^2 + \norm{W^{\ast (i)}}^2 \leqslant \frac{4}{\gamma} \sum_{i=1}^m \bar{W}_i^2 = \frac{4}{\gamma} \bar{W}$.
	
	\item By substituting $\tilde{K}_{t+1} = \bar{K}_{t+1} - W^\ast + K_{t+1} - \bar{K}_{t+1}$ in $V_a(K_{t+1})$ and defining 
		\begin{equation*}
			\begin{aligned}
				\alpha_t & \Let (K_{t+1} - \bar{K}_{t+1})\transp (K_{t+1} - \bar{K}_{t+1}) \\
				& \quad + 2 (K_{t+1} - \bar{K}_{t+1})\transp (\bar{K}_{t+1} - W^{\ast})\\
				&= -(K_{t+1} - \bar{K}_{t+1})\transp (K_{t+1} - \bar{K}_{t+1})  \\
				& \quad + 2(K_{t+1} - \bar{K}_{t+1})\transp (K_{t+1}-W^\ast),
			\end{aligned}
		\end{equation*}
		we get
		\begin{align*}
			V_a(K_{t+1}) &= \frac{1}{\gamma} \trace(\tilde{K}_{t+1}\transp \tilde{K}_{t+1}) \\
			&= \frac{1}{\gamma}\trace \left( (\bar{K}_{t+1} - W^\ast)\transp (\bar{K}_{t+1} - W^\ast)\right) +\frac{1}{\gamma} \trace \left(\alpha_t \right).
		\end{align*}
		One important property of the projection is the following \cite[(4.61)]{Ioannou_Fidan_tutorial}:
		\begin{equation}\label{e:effect_projection}
		(W^{\ast (i)} - K_t^{(i)})\transp (\bar{K}_{t}^{(i)} - K_{t}^{(i)})  \leqslant 0 \text{ for each } i = 1, \ldots , m.
		\end{equation}
		Since $(K_{t+1}^{(i)} - \bar{K}_{t+1}^{(i)})\transp (K_{t+1}^{(i)}-W^\ast) \leqslant 0$ due to \eqref{e:effect_projection}, we can ensure $\trace(\alpha_t) \leqslant 0$. Therefore,
		\begin{align*}
			& V_a(K_{t+1}) \leqslant \frac{1}{\gamma}\trace \left( (\bar{K}_{t+1} - W^\ast)\transp (\bar{K}_{t+1} - W^\ast)\right) \\
			&= V_a(K_t) + \frac{\gamma}{\norm{\phi_j(\st_t, \control_t)}^2} \trace \left( \tilde{\st}_{t+1} \tilde{\st}_{t+1}\transp \right)  \\
			&\quad - \frac{1}{\norm{\phi_j(\st_t, \control_t)}^2} \trace\left( \tilde{K}_t \transp \phi_j(\st_t, \control_t) \tilde{\st}_{t+1}\transp + \tilde{\st}_{t+1}\phi_j(\st_t, \control_t)\transp \tilde{K}_t \right).
		\end{align*}
		By substituting $\tilde{K}_t \transp \phi_j(\st_t, \control_t) = \tilde{\st}_{t+1} + \varepsilon_j(\st_t,\control_t) $ in the above inequality, we get
		\begin{align*}
			& V_a(K_{t+1})  \leqslant V_a(K_t) + \frac{1}{\norm{\phi_j(\st_t, \control_t)}^2} \biggl( \gamma \norm{\tilde{\st}_{t+1}}^2 - 2 \trace(\tilde{\st}_{t+1} \\
			& \qquad + \varepsilon_j(\st_t)) \tilde{\st}_{t+1}\transp \biggr) \\
			&= V_a(K_t) + \frac{1}{\norm{\phi_j(\st_t,\control_t)}^2} \biggl( \gamma \norm{\tilde{\st}_{t+1}}^2  \\
			& \qquad - 2 \tilde{\st}_{t+1}\transp(\tilde{\st}_{t+1} + \varepsilon_j(\st_t,\control_t))  \biggr) \\
			& = V_a(K_t) + \frac{1}{\norm{\phi_j(\st_t,\control_t)}^2} \left( (\gamma -2) \norm{\tilde{\st}_{t+1}}^2  - 2 \tilde{\st}_{t+1}\transp \varepsilon_j(\st_t,\control_t)  \right) \\
			& \leqslant V_a(K_t) + \frac{1}{\norm{\phi_j(\st_t,\control_t)}^2} \left( (\gamma -1) \norm{\tilde{\st}_{t+1}}^2  + \norm{\varepsilon_j(\st_t,\control_t)}^2  \right) \\
			& = V_a(K_t) - \frac{1- \gamma}{\norm{\phi_j(\st_t,\control_t)}^2} \norm{\tilde{\st}_{t+1}}^2 + \frac{\norm{\varepsilon_j(\st_t,\control_t)}^2}{\norm{\phi_j(\st_t,\control_t)}^2}  \\
			& \leqslant V_a(K_t) - \frac{1- \gamma}{\sigma^2} \norm{\tilde{\st}_{t+1}}^2 +  \norm{\varepsilon_j(\st_t,\control_t)}^2,
		\end{align*}	
		where the last inequality is due to $1 \leqslant \norm{\phi_j(\st_t,\control_t)}^2 \leqslant \sigma^2$. Therefore,
		\begin{equation*}
			V_a(K_{t+1}) - V_a(K_t)  \leqslant - \frac{1- \gamma}{\sigma^2} \norm{\tilde{\st}_{t+1}}^2 +  \norm{\varepsilon_j(\st_t,\control_t)}^2 .
		\end{equation*}
		\item Consider Lemma \ref{lem:adaptation}-\ref{lem:drift_Va} to get
		\begin{align*}
			\frac{1 - \gamma }{\sigma^2 }\norm{\tilde{\st}_{t+1}}^2 & \leqslant - V_a(K_{t+1}) + V_a(K_t) +  \norm{\varepsilon_j(\st_t,\control_t)}^2 \\
			& \leqslant - V_a(K_{t+1}) + V_a(K_t) + \bar{\varepsilon}^2.
		\end{align*}
		By summing from $t=k$ to $k+N-1$ in both sides, we get
		\begin{equation*}
			\begin{aligned}
				&\frac{1 - \gamma }{\sigma^2 } \sum_{t=k}^{k+N-1} \norm{\tilde{\st}_{t+1}}^2 \leqslant V_a(K_k) + N \bar{\varepsilon}^2, \\
				& \quad \leqslant \frac{4}{\gamma} \bar{W} + N \bar{\varepsilon}^2.  
			\end{aligned}
		\end{equation*}
		Therefore, $\tilde{\st}_t$ is $\bar{\varepsilon}^2$ small in mean square sense with $c_0 = \frac{\sigma^2}{1-\gamma }$ and $c_0^{\prime} = \frac{4c_0}{\gamma}\bar{W}$ as per the Definition \ref{def:small_disturbance}.	
	\end{enumerate}			 	
\end{proof}
We recall the following definition:
\begin{definition}[\cite{nonrobustMPC_examples}]
\rm{
	A system is robust asymptotically stable around $x^r$ if there exists a class-$\mathcal{KL}$ function $\beta$ and for every $\varepsilon > 0$ there exists $\delta > 0$ such that $\norm{h(\st_t, \control_t)} \leqslant \delta \implies \norm{\st_t - \st^r} \leqslant \beta(\norm{\st_t - \st^r},t) + \varepsilon$ for all $t$.
 }
\end{definition}
\begin{theorem}[Recursive feasibility and stability]
	The control law of LBMPC \eqref{e:LBMPC} is robust asymptotically stable with respect to the disturbances. The closed-loop system \eqref{e:system} under LBMPC \eqref{e:LBMPC} is input-to-state stable. The optimization problem \eqref{e:LBMPC} is recursively feasible.
\end{theorem}
\begin{proof}
	Since $\hat{h}_t$ is a continuous and uniformly bounded function due to the construction, the result follows from  \cite[Theorem 2]{lbmpc_linear}. The second result is a direct implication of \cite[Lemma 3.5]{ISS_discrete}. The recursive feasibility of \eqref{e:LBMPC} is due to \cite[Theorem 1]{lbmpc_linear}. In particular, if $(c_0,c_1, \ldots , c_{N-1})$ is an optimizer of \eqref{e:LBMPC} at some time $t$ then $(c_1, \ldots , c_{N-1}, 0)$ will be a feasible solution at time $t+1$.
\end{proof}

	\section{Numerical experiment}\label{s:experiment}
	 We consider the compression system of a jet engine, which exhibits instabilities due to rotating stall and surge. The Moore-Greitzer compressor model is given by the following nonlinear dynamics:
	 \begin{equation}\label{e:Moore-Greitzer-model}
	 \begin{aligned}
	     \dot{z} &= -y + z_c + 1 +\frac{3}{2}z - \frac{1}{2}z^3\\
	     \dot{y} &= \frac{1}{\beta^2}\left( z + 1 - r\sqrt{z} \right),
	 \end{aligned}
	 \end{equation}
	 where $z$ is mass flow, $y$ is pressure rise, $\beta>0$ is a constant and $r$ is the throttle opening. Similar to \cite{lbmpc_linear}, we assume that $r$ is controlled by a second order actuation with transfer function $r(s) = \frac{\omega_n^2}{s^2 + 2\zeta \omega_n s + \omega_n^2} \control(s)$, where $\control$ is the input. Our simulation parameters are $\beta = 1, z_c = 0, \zeta = \frac{1}{\sqrt{2}}, \omega_n = 10\sqrt{10} $. We have the following constraints:
	 \begin{equation}
	     \begin{aligned}
	          z &\in [0, 1] \quad 
	          y \in [1.1875, 2.1875]\\
	          r & \in [0.1547, 2.1547] \quad 
	          \dot{r} \in [-20, 20] \\
	          u & \in [0.1547, 2.1547]
	     \end{aligned}
	 \end{equation}
The state $x$ of the system is represented by $x = \bmat{z & y & r & \dot{r}}\transp \in \R^4$. The nonlinear model \eqref{e:Moore-Greitzer-model} is linearized around the equilibrium point $x_e = \bmat{0.5 & 1.6875 & 1.1547 & 0}\transp$ and discretized with sampling time $T=0.05$ seconds. We chose $T=0.05$ to make it slightly larger than the solver time in one optimization problem corresponding to linear MPC for its online implementation. We use a learning rate of 0.001 to train the deep neural network in parallel. 

We employed CasADI for the symbolic representation of the underlying optimization problem of MPC \eqref{e:LBMPC}. This symbolic representation is done offline and therefore, helpful in online implementation of MPC by updating only the measured state. Since PyTorch has inbuilt tools for the training of neural networks, we want to use it along with CasADI. We design a neural network on main machine with the help of CasADI and that on the secondary machine by PyTorch. 

Another difficulty is associated with the computation of terminal set $\Omega$ and reachable sets $R_i$'s. The computation of $R_i$ is very hard for large $i$. MATLAB based tools like MPT are not available in python for polytopic manipulation. The python library pytope has very limited capability and is under development. Therefore, we followed the approach of \cite{lbmpc_extension}. Let the set $\mathcal{X} \ominus \mathds{W}$, $\mathcal{X}$ and $\controlset$ be given by
		\begin{align*}
		    \mathcal{X} \ominus \mathds{W} &\Let \{x \in \R^d \mid F_p x \leqslant h_p \} \\
		    \mathcal{X} & \Let \{x  \in \R^d \mid F_x x \leqslant h_x \} \\
		    \controlset & \Let \{x \in \R^m \mid F_u x \leqslant h_u \},
		\end{align*}
where $F_p, F_x, F_u$ are suitable matrices and $h_p, h_x, h_u$ are suitable vectors required to provide the half-space representations of above sets.
The set $\Omega$ and $R_i$ are approximated in \cite{lbmpc_extension}.

\begin{figure}
    \centering
    \includegraphics[width=0.9\linewidth]{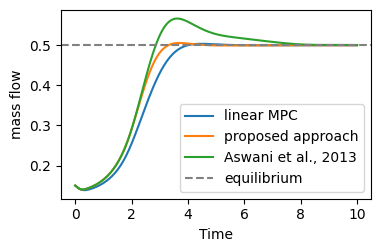}
	\caption{Proposed approach and \cite{lbmpc_linear} both have faster transient response than that of linear MPC but the proposed approach has smaller overshoot than that of \cite{lbmpc_linear}.}
	\label{fig:mass_flow}
\end{figure}

\begin{figure}
    \centering
	\includegraphics[width=0.85\linewidth]{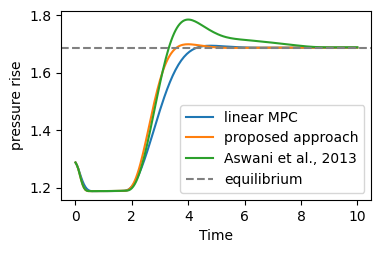}
	\caption{Proposed approach and \cite{lbmpc_linear} both have faster transient response than that of linear MPC but the proposed approach has smaller overshoot than that of \cite{lbmpc_linear}.}
	\label{fig:pressure_rise}
\end{figure}

\begin{figure}
    \centering
	\includegraphics[width=0.85\linewidth]{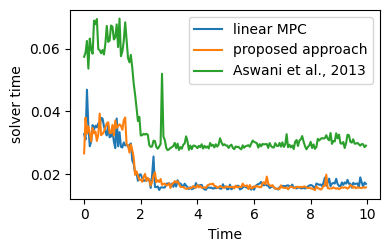}
	\caption{Solver time in both linear MPC and proposed approach is smaller than that of \cite{lbmpc_linear}.}
	\label{fig:solver_time}
\end{figure}

We are interested in comparing our proposed approach with the state-of-the-art \cite{lbmpc_linear}, which employs L2NW as an estimator. We implemented L2NW with the help of CasADI in terms of symbolic variables. One of the inputs for L2NW is a data buffer of fixed size, used as function parameters to learn the uncertainties. In an online implementation, this data buffer is empty at the beginning and its size increases with time. Therefore, we initialize a fixed size buffer in such a way that default values do not affect the outcome of the L2NW estimator. Our simulation parameters are same as in \cite{lbmpc_linear} except the sampling time $T$. Figures \ref{fig:mass_flow} and \ref{fig:pressure_rise} demonstrate that the proposed approach has faster transient response than that of linear MPC and smaller overshoot than that of \cite{lbmpc_linear}. It is demonstrated in Fig.\ \ref{fig:solver_time} that the proposed approach and linear MPC both have comparable solver time but that in \cite{lbmpc_linear} is larger than the sampling time at the beginning.  

Only drawback in \cite{lbmpc_linear} is the use of L2NW estimator, which makes the underlying optimization problem of LBMPC computationally demanding and therefore hard to implement on a small machines. Since the approach \cite{lbmpc_linear} gives choice to use any estimator, authors used linear estimators in \cite{lbmpc_quadcopter, lbmpc_extension} instead of L2NW estimator. Our approach of using DNN as an estimator is not only computationally less demanding due to the parallel processing but surprisingly the underlying optimization problems have similar solver time as linear MPC (Fig.\ \ref{fig:solver_time}). Therefore, the proposed approach can be implemented on small machines.
	 
\section{Conclusion}\label{s:epilogue}
This article demonstrates that DNN can be used in the framework of LBMPC by dual time scale training and using bounded neurons in the last activation layer. The proposed approach is able to provide a faster solution because a DNN can be trained on a separate machine. Since LBMPC allows any estimator as an oracle, different DNN architectures can be investigated for different class of problems by following the proposed approach. Some interesting extensions may be possible along the lines of vision based navigation \cite{Gasparino_RAL_2022}, stochastic MPC \cite{PDQ_intermittent, ref:PDQ-15}, Bayesian Neural Network \cite{BNN_Mesbah} and crystallization processes \cite{bal2020, bal2022}.

	\bibliographystyle{IEEEtran}
	\bibliography{MPC_Learning}

\end{document}